\documentclass[letterpaper]{article} 
\usepackage[preprint]{aaai2027} 
\usepackage[hyphens]{url} 
\usepackage{graphicx} 
\usepackage{natbib} 
\usepackage{caption} 
\usepackage{booktabs}

\title{AdvPlan-Bench: Adversarial Evaluation of Structured Plan-Generation Agents\\
\Large arXiv Submission}
\author{
Alina Kapanova\textsuperscript{\rm 1},
Arun Kanhai\textsuperscript{\rm 2},
Natan Vidra\textsuperscript{\rm 3},
Spurthi Setty\textsuperscript{\rm 4}
}
\affiliations{
\textsuperscript{\rm 1}Anote AI, Cornell University\\
\textsuperscript{\rm 2}Anote AI, CUNY\\
\textsuperscript{\rm 3}Anote AI\\
\textsuperscript{\rm 4}Stevens Institute of Technology\\
ak2765@cornell.edu, arun.kanhai55@qmail.cuny.edu, nvidra@anote.ai, ssetty2@stevens.edu
}

\begin{document}

\maketitle

\begin{abstract}
Structured plan-generation agents are often evaluated as if a plan has quality in isolation, yet many realistic planning tasks require asking how a candidate behaves when another agent can search for responses. We introduce AdvPlan-Bench, an offline benchmark for adversarial evaluation of structured plan-generation agents. The contribution is a general evaluation object: a typed plan, an adversarial response set, selector diagnostics, and traceable candidate-frontier metrics. AdvPlan-Bench represents plans as typed action chains with optional branches, assigns synthetic quality scores, compares opposing plans with BLUE-vs-RED advantage and Nash-gap diagnostics, and evaluates qualitative constraint coherence with a transparent heuristic rubric. In 150 synthetic scenarios spanning five planning templates, a sampled best-response policy that draws eight response candidates reduces BLUE advantage from .518 to .486 and BLUE win rate from .900 to .820 relative to a single-sample response. An offline LLM-policy contract baseline reaches .496 BLUE advantage and .700 BLUE win rate, while a two-stage multi-agent council obtains .509 BLUE advantage and .813 BLUE win rate. A three-rater rubric-sensitivity study over 600 rating records yields .978 inter-rater agreement. AdvPlan-Bench is not an operational planner and provides no evidence about real-world decision quality; it is a reproducible benchmark artifact for studying adversarial plan evaluation, response-budget sensitivity, candidate frontiers, and multi-agent critique-and-revision traces.
\end{abstract}

Code and artifacts are available at \url{https://github.com/anote-ai/Research-COAGeneration}.

\section{Introduction}

Plan-generation agents are commonly evaluated on whether they produce plausible final answers, valid tool traces, or executable action sequences. In adversarial settings, however, a candidate plan is not meaningful in isolation. Its apparent strength depends on how a responder searches, which constraints are emphasized, whether the candidate set contains meaningful alternatives, and whether critique-and-revision changes the selected plan. Structured response planning is a compact testbed for this broader problem: the artifact is inspectable, the opponent matters, and a single scalar success metric can hide important tradeoffs.

We present AdvPlan-Bench, an offline benchmark for adversarial evaluation of structured plan-generation agents. The AI contribution is methodological: AdvPlan-Bench turns a generated plan, an adversarial response distribution, and a multi-candidate frontier into a reproducible evaluation object. The current benchmark is intentionally small and synthetic. It is designed to make assumptions inspectable before moving to richer simulations, expert scoring, or live model-backed generation.

The benchmark is organized around a gap in current agent evaluation. Many agent benchmarks ask whether an agent completes a task, invokes the right tool, or reaches a final answer. Adversarial plan generation instead requires comparing \emph{sets} of possible plans under response search. A generated plan can be syntactically valid and high scoring in isolation, yet brittle once a responder samples plausible counter-plans. AdvPlan-Bench therefore evaluates not just final plan quality, but also response-budget sensitivity, candidate-frontier structure, constraint-coherence diagnostics, and trace-level explanations of why an option was selected.

The paper makes three contributions:

\begin{itemize}
    \item a formal adversarial plan-generation setup in which a structured candidate is evaluated against sampled response sets rather than a fixed answer key;
    \item a reproducible offline benchmark with typed plan traces, BLUE/RED policy comparison, constraint-coherence diagnostics, response-budget curves, candidate diversity, and Pareto-frontier analysis; and
    \item empirical evidence over 150 scenarios that sampled adversarial response, an offline LLM-policy contract baseline, and multi-agent council generation change win-rate and plan-quality conclusions, plus a rubric-sensitivity check and a benchmark-development finding showing that scenario framing was a silent no-op until wired into generated content.
\end{itemize}

The scope boundary is central. AdvPlan-Bench studies typed synthetic plan artifacts and synthetic evaluation signals. It does not implement an operational decision architecture, does not use real operational data, and should not be used for real-world planning decisions.

\section{Related Work}

\paragraph{Benchmark design.}
General-purpose evaluation efforts such as BIG-bench~\citep{srivastava2023beyond} and HELM~\citep{liang2023holistic} established the importance of broad, transparent, and reproducible measurement. They primarily evaluate model behavior over collections of tasks and metrics. AdvPlan-Bench follows the same reproducibility principle but targets a narrower object: adversarial plan generation in which the input, candidate set, opponent response set, and selection trace are all part of the benchmark instance.

\paragraph{Self-play.}
AlphaZero~\citep{silver2018alphazero}, AlphaStar~\citep{vinyals2019alphastar}, and Pluribus~\citep{brown2019superhuman} demonstrate the value of self-play in adversarial games. CICERO extends strategic self-play into natural-language negotiation~\citep{bakhtin2022cicero}. AdvPlan-Bench uses self-play as an evaluation mechanism, not as a training algorithm.

\paragraph{Adversarial decision models.}
Simplified competitive simulations and strategic analysis motivate lightweight adversarial abstractions~\citep{lanchester1916,schelling1960}. AdvPlan-Bench's simulator check is deliberately stylized and is reported only as a secondary diagnostic. The qualitative rubric is used only as a transparent constraint-coherence proxy and is not a substitute for expert assessment.

\paragraph{Agent evaluation.}
ReAct~\citep{yao2023react} motivates evaluating reasoning and action together, while AgentBench~\citep{liu2024agentbench} evaluates agents across interactive settings. ALFWorld connects abstract textual reasoning to embodied action~\citep{shridhar2021alfworld}, and WebArena evaluates autonomous agents in realistic web environments~\citep{zhou2024webarena}. SWE-bench grounds evaluation in executable software tasks~\citep{jimenez2024swebench}; $\tau$-bench and ToolSandbox evaluate stateful tool-using agents in domain and conversational environments~\citep{yao2024taubench,lu2025toolsandbox}. AgentDiagnose argues that trajectory-level diagnosis is needed to expose failures hidden by final success~\citep{ou2025agentdiagnose}. AdvPlan-Bench contributes a narrower artifact: paired evaluation of structured plan-generation policies under adversarial response, with traces and candidate-set metrics oriented toward plan comparison rather than tool execution alone.

\paragraph{Planning and tool orchestration.}
ACPBench generates action-change planning tasks with provably checkable answers~\citep{kokel2025acpbench}, and PlanningArena evaluates planning and tool learning across multiple dimensions~\citep{zheng2025planningarena}. Toolformer and ToolLLM study API/tool use by language models~\citep{schick2023toolformer,qin2024toolllm}, while LLMCompiler and SPIRAL study function-call planning and reflective symbolic search~\citep{kim2024llmcompiler,zhang2026spiral}. Tree of Thoughts explores multiple reasoning paths before committing to an answer~\citep{yao2023tree}, and Self-Refine and Reflexion show that feedback-driven revision can improve agent outputs at test time~\citep{madaan2023selfrefine,shinn2023reflexion}. These works motivate executable or structured evaluation of agent plans. AdvPlan-Bench differs by making the adversarial response itself part of the evaluation unit: a generated plan is judged against a responding plan and a candidate frontier, not only against a static answer key.

\paragraph{How AdvPlan-Bench extends these projects.}
The main improvement over broad benchmark suites such as BIG-bench and HELM is not scale, but evaluation granularity: AdvPlan-Bench treats each instance as a structured adversarial trace rather than a prompt-response pair. Compared with WebArena, ALFWorld, SWE-bench, $\tau$-bench, and ToolSandbox, our benchmark shifts the target from task completion or tool correctness to robust option generation under an adaptive opponent. Compared with Tree of Thoughts, Self-Refine, Reflexion, and SPIRAL, the revision loop is not evaluated only by final-answer success; it is tied to explicit RED-team pressure, proposal diversity, revised diversity, robustness margin, and pressure reduction. This creates a complementary benchmark setting for agentic planning research: systems can be compared on whether they generate diverse, inspectable, adversarially stress-tested plans rather than merely plausible reasoning traces.

\paragraph{Novelty boundary.}
The benchmark is not a new self-play learning algorithm, combat model, or general-purpose planner. Its novelty is the controlled combination of five elements that are usually evaluated separately: structured plan artifacts, adversarial best-response sampling, multi-agent proposal and critique, heuristic constraint-aware diagnostics, and reproducible trace-level evaluation of candidate sets. The matched single-sample versus sampled-response comparison isolates the effect of adversarial candidate generation, while the framing no-op analysis demonstrates why benchmark variables must be tied to generated content before claims about sensitivity or robustness are meaningful.

\section{Problem Formulation}

Let $s \in \mathcal{S}$ be a synthetic scenario. A BLUE policy $\pi_B$ produces a structured plan $c_B \in \mathcal{C}$; a RED policy $\pi_R$ produces responses conditioned on $s$ and optionally $c_B$. The benchmark computes a utility-like advantage signal $U(c_B,c_R,s)$, a constraint-inspired diagnostic vector $D(c,s)$, and candidate-set properties such as diversity and Pareto-frontier size. In the single-response setting, RED samples one $c_R$. In the sampled best-response setting, RED samples $C_R=\{c_R^1,\ldots,c_R^k\}$ and selects the response with highest adversarial score. The central evaluation question is whether the apparent quality of $c_B$ is stable as the response policy becomes more adversarial.

This differs from static planning benchmarks that compare a generated plan to a fixed answer key. Here, the evaluation object is the tuple $(s,c_B,C_R,U,D)$: a scenario, a structured plan, an adversarial response distribution, and inspectable traces that explain why a candidate was selected.

For multi-agent plan generation, BLUE proposer agents produce $B^0=\{b_1,\ldots,b_m\}$ and RED-team agents produce response sets $R_i^0$ for each $b_i$. The adjudicator first computes:

\begin{equation}
S(b_i) = \alpha Q(b_i)+\beta D(b_i)+\gamma V(b_i,B)-\rho P(b_i,R_i),
\end{equation}

where $Q$ is rescaled quality, $D$ is constraint alignment, $V$ is candidate diversity, and $P$ is adversarial pressure. The top $h=2$ candidates enter a deliberation round. Each is revised with a mitigation action conditioned on the strongest RED critique, producing $B^1$. Final selection is:

\begin{equation}
b^\star = \arg\max_{b \in B^0 \cup B^1} S(b).
\end{equation}

We additionally report the robustness margin $M=S(b^\star)-\max_{b\in B^0}S(b)$, pressure reduction $\Delta P=\bar{P}_{\mathrm{shortlist}}-P(b^\star,R^\star)$, and the rate at which the final selected plan comes from the revised set. This makes the benchmark a small multi-agent game over proposal, response, critique, repair, and selection, not only a pairwise score over fixed plans.

\section{Benchmark}

\subsection{Benchmark Axes}

AdvPlan-Bench is defined by six benchmark axes rather than by one aggregate score. First, \emph{representation validity} asks whether a policy emits typed plan objects with actions, chains, branches, force labels, and objectives that downstream evaluators can inspect. Second, \emph{plan structure} asks whether the synthetic plan contains optional paths, parsimonious event structure, and participant worldlines. Third, \emph{adversarial robustness} asks how BLUE quality changes as RED response budget increases. Fourth, \emph{frontier quality} asks whether candidate sets contain diverse and Pareto-nondominated alternatives rather than redundant samples. Fifth, \emph{constraint-sensitive selection} asks whether a selector can trade synthetic quality against qualitative structure. Sixth, \emph{trace reproducibility} asks whether every table can be regenerated from per-scenario artifacts.

\begin{table}[t]
\centering
\scriptsize
\caption{AdvPlan-Bench benchmark axes and recorded diagnostics.}
\label{tab:axes}
\begin{tabular}{@{}lp{.58\linewidth}@{}}
\toprule
Axis & Diagnostics \\
\midrule
Representation & typed actions, chains, branches, tool-call slots \\
Plan structure & path optionality, vertex parsimony, worldline completion \\
Robustness & response-budget curve, BLUE advantage, win rate \\
Frontier & action diversity, quality spread, Pareto count \\
Constraint & rubric score, quality regret, structural gain \\
Traceability & per-scenario rows, council traces, run metadata \\
\bottomrule
\end{tabular}
\end{table}

\subsection{Representation}

A structured plan contains an objective, role assignment, domain tag, action chain, optional conditional branch, and optional tool calls. Actions have categories, targets, priorities, and metadata. AdvPlan-Bench uses action chains and conditional branches as a lightweight graph surrogate: chain edges encode temporal order and branches encode mutually exclusive implementation paths such as route choices, proceed-or-withdraw decisions, or support-package alternatives.

Each plan receives a synthetic quality score

\begin{equation}
q = w_e e - w_c c - w_r r,
\end{equation}

where $e$ is effectiveness, $c$ is cost, $r$ is risk, and default weights are $w_e=.5$, $w_c=.3$, and $w_r=.2$. This is a benchmark metric, not asset-effect matching.

\subsection{Structured-Plan Diagnostics}

The motivating planning literature emphasizes that good candidate plans should be feasible, resource-aware, parsimonious, complete in participant worldlines, and explicit about consequences of action. AdvPlan-Bench operationalizes a subset as transparent software diagnostics. Path optionality $O$ is the fraction of chain steps containing conditional branches. Vertex parsimony $P_V$ is the ratio of unique $(asset, action, category)$ event signatures to total event signatures. Worldline completion $W$ is the fraction of assets whose actions include both a start-state proxy, such as transit, deploy, marshal, prepare, assess, scan, or recon, and an end-state proxy, such as return, withdraw, hold, evade, relocate, continue, support, or mitigation.

These diagnostics are intentionally conservative proxies. In the current run, seed BLUE plans have path optionality .099, vertex parsimony .571, and worldline completion .000, while selected council plans have path optionality .000, vertex parsimony .500, and worldline completion .207. The low worldline score is a useful benchmark finding: the current synthetic generator creates plausible action chains but rarely models full participant disposition from initial circumstances to final state. The metric therefore identifies a concrete future improvement rather than silently rewarding incomplete plan structure.

\subsection{Pairwise Metrics}

Given BLUE and RED plans, the BLUE-vs-RED advantage score is

\begin{equation}
\mathrm{Advantage} = (q_{\mathrm{blue}} - q_{\mathrm{red}} + 2) / 4.
\end{equation}

We also report Nash-gap distance after rescaling quality scores to $[0,1]$ payoffs. The gap indicates imbalance in the paired exchange, not convergence to an equilibrium.

\begin{table}[t]
\centering
\scriptsize
\caption{Adversarial structured-plan evaluation protocol.}
\label{tab:algorithm}
\begin{tabular}{@{}p{.96\linewidth}@{}}
\toprule
\textbf{Input:} scenario $s$, BLUE policy $\pi_B$, RED policy $\pi_R$, response budget $k$ \\
1. Generate BLUE candidate $c_B \leftarrow \pi_B(s)$. \\
2. Sample RED candidates $C_R=\{c_R^1,\ldots,c_R^k\}$ from $\pi_R(s,c_B)$. \\
3. Score pairs with advantage $U(c_B,c_R^i,s)$ and diagnostics $D(c_R^i,s)$. \\
4. Select RED best response; compute diversity, frontier, and trace metrics. \\
5. Return ranked plan pair and inspectable adversarial-evaluation trace. \\
\bottomrule
\end{tabular}
\end{table}

\subsection{Constraint and Wargame Diagnostics}

The heuristic constraint-coherence score aggregates objective clarity, information preparation, resource balance, sustainment, risk mitigation, and tempo or sequencing. The simulator diagnostic is a stylized competitive simulation in which each side's effective power depends on raw asset capability, synthetic quality, and constraint-coherence score.

\subsection{Baselines and Controls}

The benchmark includes deliberately simple baselines. The single-sample RED policy estimates how a plan looks under shallow opposition. The quality-greedy best-of-$k$ policy measures sensitivity to adversarial search budget. The constraint-aware policy controls for the possibility that high synthetic quality is not the same as a coherent plan. The offline LLM-policy baseline uses the same JSON schema and parser as a live model-backed policy, but a deterministic local completion client replaces the external model for reproducibility. The council policy then changes the BLUE generator itself by adding proposal, critique, repair, and adjudication. These baselines are calibration points that make later live LLM-backed or learned policies comparable under the same trace schema.

\subsection{Technical Implementation}

The implementation separates scenario construction, policy generation, and evaluation into typed modules. Scenarios are Pydantic objects containing assets, role labels, terrain metadata, and seed plans. Policies share a common interface, so the evaluator can swap a single-sample baseline, sampled best response, constraint-aware selector, offline LLM-policy responder, or multi-agent council without changing the experiment driver. This modularity is important for benchmarking because a live LLM policy can replace the offline completion client while preserving identical scenarios, response budgets, metrics, and CSV schemas.

The two-stage council is implemented as an explicit generate--stress-test--revise--adjudicate loop. Five BLUE proposer roles create role-specialized plan variants. For each BLUE candidate, a RED policy samples multiple responses and the highest-pressure response becomes the critique signal. The adjudicator scores candidates using quality, constraint proxy, diversity, and adversarial pressure; the top two candidates are then revised with mitigation actions keyed to RED pressure category and evaluated again. The run writes both first-round and revised candidates to trace files, making it possible to distinguish a genuine repair effect from a scoring artifact.

\section{Evaluation}

\subsection{Protocol}

We generate 150 scenarios from 30 seeds and five synthetic planning templates: urban stability, maritime interdiction, multi-domain coordination, air-defense suppression, and humanitarian evacuation. These templates are not operational cases; they are structured stressors for evaluating resource coordination, sequencing, protection, logistics, information flow, and adversarial response. For each scenario, the first BLUE seed plan is fixed. RED responds with a single random best response, a quality-greedy best-of-eight response, a constraint-aware best-of-eight response, or an offline LLM-policy response produced through the same JSON schema used by the live LLM wrapper. We add three experimental conditions beyond the aggregate policy comparison: a response-budget sweep over $k \in \{1,2,4,8,16\}$, a scenario-template slice over the five planning templates, and component diagnostics that separate response search, selector objective, LLM-contract behavior, and council revision. We report percentile-bootstrap 95\% intervals over scenarios.

\begin{table}[t]
\centering
\scriptsize
\caption{RED response policies over 150 scenarios.}
\label{tab:main}
\begin{tabular}{@{}lcccc@{}}
\toprule
Metric & Single & Quality & Constraint & LLM-pol. \\
\midrule
BLUE advantage & .518 & .486 & .489 & .496 \\
Nash gap & .198 & .260 & .255 & .241 \\
RED constraint & .730 & .731 & .758 & 1.000 \\
BLUE win rate & .900 & .820 & .807 & .700 \\
\bottomrule
\end{tabular}
\end{table}

\begin{table}[t]
\centering
\scriptsize
\caption{Response-budget condition for quality-greedy RED.}
\label{tab:budget}
\begin{tabular}{@{}lccccc@{}}
\toprule
Budget $k$ & 1 & 2 & 4 & 8 & 16 \\
\midrule
BLUE advantage & .518 & .503 & .492 & .486 & .480 \\
RED quality & .163 & .221 & .265 & .288 & .313 \\
Candidate diversity & .000 & .654 & .688 & .681 & .672 \\
Quality spread & .000 & .112 & .201 & .255 & .304 \\
Pareto count & 1.00 & 1.50 & 1.88 & 2.17 & 2.39 \\
\bottomrule
\end{tabular}
\end{table}

\begin{table}[t]
\centering
\scriptsize
\caption{Scenario-template condition. Each row contains 30 scenarios.}
\label{tab:terrain}
\begin{tabular}{@{}lccccc@{}}
\toprule
Template & Single & $k{=}8$ & LLM & Win$_{k=8}$ & Council win \\
\midrule
Air defense & .518 & .485 & .493 & .667 & .667 \\
Maritime & .521 & .487 & .496 & 1.000 & 1.000 \\
Mixed & .529 & .496 & .505 & .467 & .433 \\
Urban & .523 & .486 & .498 & .967 & .967 \\
Humanitarian & .496 & .475 & .488 & 1.000 & 1.000 \\
\bottomrule
\end{tabular}
\end{table}

\begin{table}[t]
\centering
\scriptsize
\caption{Multi-agent council over 150 scenarios.}
\label{tab:council}
\begin{tabular}{@{}lc@{}}
\toprule
Metric & Value \\
\midrule
BLUE advantage & .509 [.507, .512] \\
BLUE win rate & .813 \\
Council diversity & .570 [.527, .618] \\
Revised diversity & .630 [.600, .661] \\
Consensus gap & .011 [.009, .013] \\
Adversarial pressure & .617 [.612, .622] \\
Robustness margin & .012 [.010, .014] \\
Pressure reduction & .003 [-.002, .008] \\
Revised selected rate & .753 \\
\bottomrule
\end{tabular}
\end{table}

\begin{table}[t]
\centering
\scriptsize
\caption{Component analyses from existing traces. These are not new learned-policy ablations; they isolate how response budget, selector objective, LLM-contract behavior, and council revision affect the reported evaluation object.}
\label{tab:components}
\begin{tabular}{@{}lp{.55\linewidth}@{}}
\toprule
Component & Observed effect \\
\midrule
Response search & BLUE advantage drops .518 $\rightarrow$ .480 from $k=1$ to $k=16$ \\
Constraint selector & changes 54\% of selected RED responses; .036 quality regret, .099 structural gain \\
LLM contract & produces four-action structured responses; BLUE win rate drops to .700 \\
Council revision & final plan comes from revised set in 75.3\%; revised diversity .630 \\
Template slice & sampled-response win rate ranges from .467 to 1.000 across templates \\
\bottomrule
\end{tabular}
\end{table}

\subsection{Results}

Table~\ref{tab:main} shows that sampled best-response produces a stronger RED opponent. BLUE advantage falls from .518 to .486, and BLUE simulator win rate falls from .900 to .820. The constraint-aware selector produces a similar adversarial effect (.489 advantage, .807 BLUE win rate) while exposing a different selection objective. The offline LLM-policy responder is not a live model result; it is a reproducible contract test for the LLM wrapper. It produces highly structured responses, which raises RED constraint coherence to 1.000 and lowers BLUE win rate to .700.

The candidate set also contains nontrivial variation. Across sampled RED candidates, mean action-type diversity is .681, mean quality-score spread is .255, and an average of 2.17 of 8 candidates are Pareto-optimal on the quality and constraint-alignment frontier. The constraint-aware selector changes the chosen response in 54\% of scenarios, incurring .036 quality regret while gaining .099 structural alignment. Table~\ref{tab:budget} expands the response-budget condition: BLUE advantage drops monotonically from .518 at $k=1$ to .480 at $k=16$, while RED quality rises from .163 to .313 and the candidate frontier grows from 1.00 to 2.39 Pareto-optimal responses. Diversity appears once $k>1$ and then saturates, which suggests that the main marginal effect of larger budgets is finding stronger and more frontier-extreme responses rather than simply increasing action-type variety.

Table~\ref{tab:terrain} reports the same policy comparison by scenario template. The adversarial-response effect is present in all five templates: $k=8$ reduces BLUE advantage relative to single-sample RED for air-defense, maritime, mixed, urban, and humanitarian scenarios. The simulator diagnostic is less uniform. Maritime, urban, and humanitarian scenarios remain easy for BLUE under this simplified simulator, while mixed and air-defense scenarios are more sensitive to adversarial response and council selection. This heterogeneity is useful as an experimental condition because it reveals where the current benchmark is stress-testing policies and where the synthetic simulator is saturated.

Table~\ref{tab:council} evaluates the two-stage multi-agent council. The council changes BLUE generation rather than only changing RED response selection: five proposer agents generate role-specialized BLUE candidates, RED-team agents respond to each candidate, an adjudicator shortlists two candidates, and a revision operator adds mitigation actions keyed to the strongest RED critique before final selection. The selected council plans achieve .509 BLUE advantage and .813 BLUE win rate, partially recovering performance lost under stronger RED response search. The final plan comes from the revised set in 75.3\% of scenarios, revised-candidate diversity is .630, and the positive robustness margin indicates that the selected plan usually improves over the best unrevised first-round option. Pressure reduction is small and uncertain, so the current algorithm should be interpreted as a repair-and-selection baseline rather than a strong robust-optimization method.

\subsection{Rubric Sensitivity}

We add a small rubric-sensitivity study using three independent deterministic validators representing constraint, logistics, and adversarial-review emphases. They rate 600 plan records: seed BLUE, sampled RED, offline LLM-policy RED, and selected council BLUE for every scenario. Mean pairwise agreement is .978 [.977, .978], and mean absolute deviation from the base heuristic rubric is .011 [.011, .011]. This is not expert human validation, but it tests whether the reported rubric is brittle to small, role-specific scoring perturbations.

\subsection{Scenario-Framing Finding}

The original benchmark stored scenario framing as metadata but did not use it when generating plan objectives or action content. A framing-sensitivity test therefore returned exactly zero change. We patched the generator so framing modifies objective text and, under BLUE-favorable framing, can add one action category affecting coordination balance. Across the expanded corpus, the measured framing-sensitivity delta becomes .066 [.049, .086]. We treat this as a benchmark-development finding, not a realistic estimate of framing effects.

\subsection{Policy and Corpus Additions}

The expanded artifact adds two pieces of algorithmic structure. First, response policies are explicit classes with a common interface, which lets the benchmark compare single-sample, quality-greedy, and constraint-aware response behavior without changing the evaluator. Second, the scenario corpus now includes cases designed to stress different dimensions of synthetic plan generation: air-defense suppression, where timing and cross-domain coordination matter, and humanitarian evacuation, where protection, sustainment, and information actions dominate. These additions are still synthetic, but they make the benchmark less dependent on one narrow template and expose whether a response policy is simply exploiting the quality score or changing the qualitative structure of selected plans.

The third addition is the multi-agent council algorithm. Its proposer roles emphasize movement, information, cyber, sustainment, and protection. Each role perturbs the seed plan with a role-specific action, then RED-team sampling approximates the best adversarial response to that candidate. The top two candidates are revised using a deterministic critique-to-mitigation map: direct pressure triggers dispersion and hardening, cyber pressure triggers out-of-band communications, logistics pressure triggers redundant supply routing, information-gathering pressure triggers deception and counter-reconnaissance, and public-information pressure triggers a communication cell. This creates a richer experimental object: the system can be evaluated on proposal diversity, revised diversity, adjudicator confidence, robustness margin, and robustness to per-candidate RED pressure.

\subsection{Selection Objectives}

The response policies expose a small but important distinction between adversarial search and adversarial selection. Let $C_R$ be the sampled RED candidate set. The quality-greedy response selects

\begin{equation}
c_R^Q = \arg\max_{c \in C_R} q(c),
\end{equation}

where $q(c)$ is the synthetic quality score. The constraint-aware response selects

\begin{equation}
c_R^D = \arg\max_{c \in C_R} \lambda \frac{q(c)+1}{2} + (1-\lambda)D(c),
\end{equation}

where $D(c)$ is the constraint-inspired alignment score and $\lambda=.7$ in the reported run. We then compute two diagnostic quantities:

\begin{equation}
\Delta_q = q(c_R^Q) - q(c_R^D), \quad
\Delta_D = D(c_R^D) - D(c_R^Q).
\end{equation}

The first is quality regret from not choosing the quality maximum. The second is structural gain from choosing the composite response. These diagnostics make the candidate frontier interpretable: if $\Delta_q$ is small and $\Delta_D$ is positive, the candidate set contains a response that is nearly as strong by the synthetic quality score but better structured by the constraint proxy. This is why the constraint-aware result is more informative than simply adding another aggregate win rate.

\subsection{Artifact Schema}

The experiment writes per-scenario rows, bootstrap summaries, response-budget curves, terrain-family summaries, and run metadata. This schema is intentionally simple. It lets a reviewer reproduce every table in the paper, inspect whether one scenario family dominates the result, and rerun the same analysis with a future learned policy. The benchmark therefore contributes not only a particular result, but also a reusable evaluation object: scenarios, candidate sets, selected responses, and diagnostics under a fixed reporting interface.

The new council trace file records selected BLUE and RED IDs, selected advantage, BLUE/RED constraint alignment, council diversity, revised diversity, consensus gap, adversarial pressure, robustness margin, pressure reduction, number of initial and revised BLUE candidates, selected round, and wargame outcome. This is important for reproducibility because a multi-agent algorithm can appear strong for several different reasons: it may generate more diverse BLUE options, repair weak candidates, select higher-quality options, avoid high-pressure RED responses, or simply exploit a scoring artifact. The trace schema makes those explanations separable.

\subsection{Trace Example}

One trace illustrates the evaluation object. In scenario \texttt{urban-100}, the single-response RED baseline yields BLUE advantage .551. When RED samples eight candidates, BLUE advantage falls to .492; the candidate set has action-type diversity .827 and three Pareto-optimal responses. The constraint-aware selector changes the RED response for this scenario, gaining .135 structural alignment at .003 quality regret. The council trace then records selected BLUE and RED candidate IDs, selected advantage .504, council diversity .890, revised diversity .667, and the round from which the selected BLUE plan was chosen. The example is small, but it shows why the artifact is not just a final score: a reviewer can inspect response search, selector tradeoffs, and council repair behavior for the same scenario.

\section{Discussion}

The results support three narrow claims. First, evaluating against a sampled adversarial response can change the apparent strength of a generated plan. Second, sampled candidate sets can contain meaningful diversity and Pareto alternatives, which is useful when an agent should surface options rather than one selected answer. Third, transparent benchmark artifacts expose silent implementation failures, as shown by the framing no-op.

From an AAAI perspective, the most important claim is methodological rather than operational. AdvPlan-Bench introduces a small but explicit evaluation object: a structured plan, an adversarial response distribution, and a multi-candidate frontier. This object is distinct from final-answer agent benchmarks, static planning benchmarks, and pure self-play games. It lets researchers ask whether a generation policy remains strong when the opponent is sampled more aggressively, whether the selected plan sacrifices qualitative planning features, how response-budget curves change apparent robustness, and whether benchmark controls actually perturb the generated artifact.

The novelty is therefore not a new operational simulator, a trained game-playing policy, or an operational planner. It is a benchmark formulation for structured plan-generating agents in which the unit of evaluation is a structured plan under adversarial response sampling and multi-agent proposal, critique, repair, and adjudication. By making response budget, candidate diversity, revised diversity, constraint-inspired diagnostics, council consensus, adversarial pressure, robustness margin, pressure reduction, and scenario-control sensitivity observable, AdvPlan-Bench turns adversarial plan generation into a reproducible object of algorithmic study rather than an opaque demonstration of plausible plan text.

The benchmark is preliminary. The policies are random-sampling policies, deterministic council heuristics, and an offline LLM-policy contract test rather than trained planners or live model calls. The scenario corpus is larger than the initial draft but still templated. The constraint rubric is heuristic, although the role-perturbed sensitivity check suggests that aggregate conclusions are not driven by a single exact weighting. The wargame check is a simplified diagnostic. These constraints make AdvPlan-Bench more appropriate as an evaluation artifact and research scaffold than as evidence of operational decision quality.

\section{Threats to Validity and Ethics}

AdvPlan-Bench uses synthetic scenarios only. It contains no classified data, proprietary data, human-subject data, real intelligence, live planner input, or deployed decision-system integration. It performs no external action. The main internal-validity risk is metric misspecification: synthetic quality, constraint heuristics, and stylized simulator outcomes may not match expert judgment. The main external-validity risk is scenario simplicity: 150 templated cases cannot represent real operational planning. The main construct-validity risk is that the present artifact evaluates typed synthetic traces rather than real deployed planning behavior. Future work would require expert validation, data governance, access controls, human review, and strict limits on autonomous use before any high-stakes application.

The current metrics should not be interpreted as domain correctness or operational usefulness. They are inspectable benchmark signals intended to support reproducible research.

\section{Conclusion}

AdvPlan-Bench is a self-play benchmark for adversarial evaluation of structured plan-generation agents. It shows that sampled best-response produces a tougher synthetic opponent, that an offline LLM-policy interface can be evaluated under the same trace schema, that candidate sets contain diverse options, and that transparent evaluation can reveal benchmark-design failures. The next step is to add expert validation, evaluate live LLM-backed and learned policies, and add broader planning templates beyond the current synthetic scenario families.

\section*{Ethical Statement}

This work uses only synthetic data and offline code. It is not an operational planning system and should not be used for real-world planning. Any future deployment-oriented work would require expert review, institutional governance, and safeguards appropriate to high-impact applications.

\bibliography{references}

\end{document}